\documentclass[final,5p,times,twocolumn]{elsarticle}

\journal{Neurocomputing}

\usepackage[dvipsnames]{xcolor}
\usepackage{multirow}

\usepackage{amssymb}
\usepackage{latexsym}
\usepackage{algpseudocode}
\usepackage{blindtext}
\usepackage{algorithm}
\usepackage{times}
\usepackage{epsfig}
\usepackage{graphicx}
\usepackage{amsmath}
\usepackage{textcase}
\usepackage{comment}
\usepackage{breqn}
\usepackage{ragged2e}
\usepackage[normalem]{ulem}
\usepackage{caption, booktabs}

\usepackage{upgreek}
\usepackage{array}
\usepackage{textcomp}
\newcommand*\rot{\rotatebox{90}}
\newcolumntype{P}[1]{>{\centering\arraybackslash}p{#1}}
\useunder{\uline}{\ul}{}
\usepackage{balance}
\usepackage{lastpage}
\usepackage[pagebackref=true,breaklinks=true,letterpaper=true,colorlinks,bookmarks=false]{hyperref}

\usepackage{todonotes}
\usepackage{xparse}
\usepackage{array}
\newcolumntype{L}[1]{>{\raggedright\let\newline\\\arraybackslash\hspace{0pt}}m{#1}}
\newcolumntype{C}[1]{>{\centering\let\newline\\\arraybackslash\hspace{0pt}}m{#1}}
\newcolumntype{R}[1]{>{\raggedleft\let\newline\\\arraybackslash\hspace{0pt}}m{#1}}
\newcolumntype{J}[1]{>{\justifying\let\newline\\\arraybackslash\hspace{0pt}}m{#1}}

\usepackage{enumitem}
\usepackage{float}

\begin{document}
\begin{frontmatter}

\title{FogAdapt: Self-Supervised Domain Adaptation for Semantic Segmentation of Foggy Images}

\author{Javed~Iqbal\corref{cor}}
\cortext[cor]{Corresponding author}
\ead{javed.iqbal@itu.edu.pk}

\author{Rehan~Hafiz}
\ead{rehan.hafiz@itu.edu.pk}

\author{Mohsen~Ali}
\ead{mohsen.ali@itu.edu.pk}

\address{Information Technology University, Pakistan}



\begin{abstract}

This paper presents FogAdapt, a novel approach for domain adaptation of semantic segmentation for dense foggy scenes. Although significant research has been directed to reduce the domain shift in semantic segmentation, adaptation to scenes with adverse weather conditions remains an open question. Large variations in the visibility of the scene due to weather conditions, such as fog, smog, and haze, exacerbate the domain shift, thus making unsupervised adaptation in such scenarios challenging. We propose a self-entropy and multi-scale information augmented self-supervised domain adaptation method (FogAdapt) to minimize the domain shift in foggy scenes segmentation. Supported by the empirical evidence that an increase in fog density results in high self-entropy for segmentation probabilities, we introduce a self-entropy based loss function to guide the adaptation method. Furthermore, inferences obtained at different image scales are combined and weighted by the uncertainty to generate scale-invariant pseudo-labels for the target domain. These scale-invariant pseudo-labels are robust to visibility and scale variations. We evaluate the proposed model on real clear-weather scenes to real foggy scenes adaptation and synthetic non-foggy images to real foggy scenes adaptation scenarios. Our experiments demonstrate that FogAdapt significantly outperforms the current state-of-the-art in semantic segmentation of foggy images. Specifically, by considering the standard settings compared to state-of-the-art (SOTA) methods, FogAdapt gains 3.8\% on Foggy Zurich, 6.0\% on Foggy Driving-dense, and 3.6\% on Foggy Driving in mIoU when adapted from Cityscapes to Foggy Zurich.

\end{abstract}


\end{frontmatter}


\section{Introduction}
\label{intro}

Semantic segmentation is one of the important components of autonomous systems, i.e., self-driving cars \cite{kitty2012we}. 
Deep Learning approaches for semantic segmentation, relying on the large tagged datasets \cite{Cordts2016Cityscapes, Ros_2016_CVPR}, have resulted in substantially improved performance \cite{sakaridis2018semantic, zhao2017pyramid, chen2018deeplab} in the last few years.
However, similar to many other supervised learning problems \cite{khodabandeh2019robust, marsde2018people}, semantic segmentation models exhibit large generalization errors \cite{zou2018unsupervised, vu2019advent, LSE_2020_Naseer}.
This behavior is ascribed to the domain shift between the distribution of the test and training data domains \cite{curr2017_ICCV, Lian_2019_pycda}. 
One challenging case of the domain adaptation can be attributed to weather conditions such as rain, fog, snowfall, lightning, and strong wind \cite{chen2019gated, sakaridis2018semantic, vachmanus2020semantic}. 
\textcolor{black}{Fog specifically degrades the visibility and contrast significantly \cite{narasimhan2003contrast,tan2008visibility}, 
deteriorating the performance of the computer vision applications, e.g., segmentation. 
Domain adaptation algorithms have been presented to overcome the domain shift (synthetic to real \cite{tsai2018learning, zou2018unsupervised, mlsl2020} or real to real \cite{chen2017no} datasets)  specific to the case of semantic segmentation.}
However, very little attention has been devoted to address domain shift caused by foggy weather conditions \cite{sakaridis2018model, sakaridis2018semantic, dai2019curriculum}.


\begin{figure*}[t]
 	\centering
 	\includegraphics[width=\textwidth]{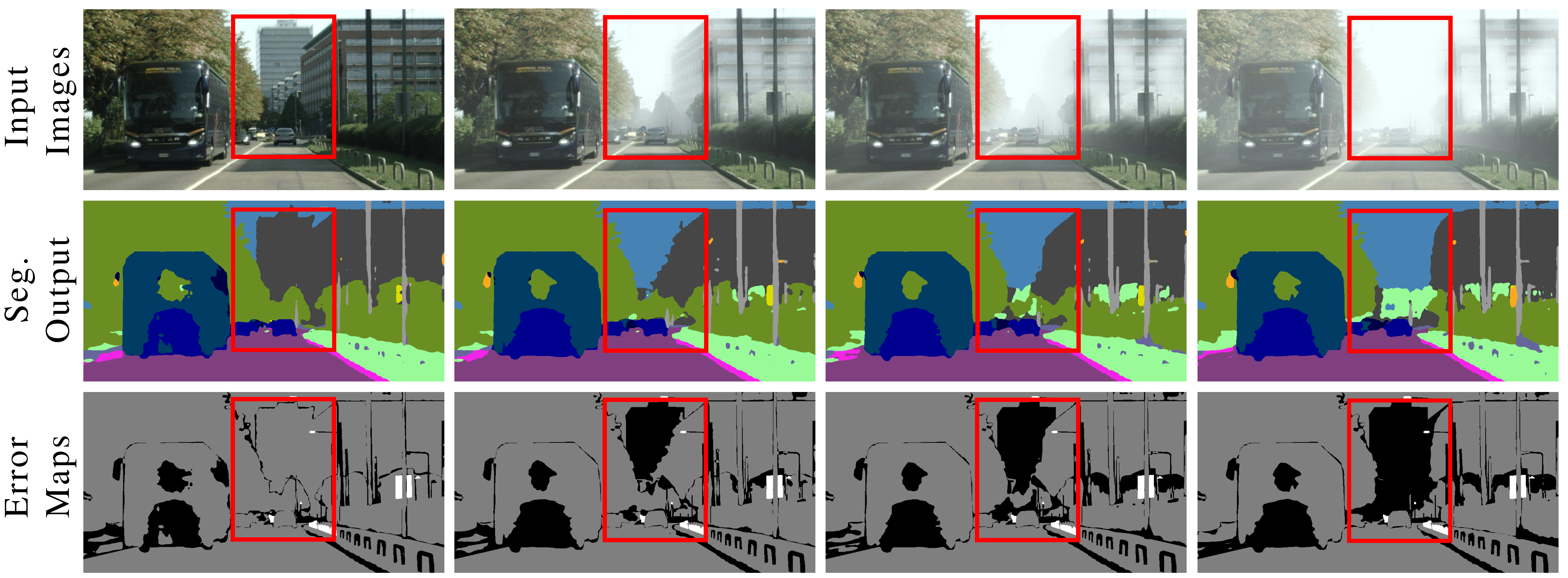}
 	\scriptsize
    \begin{tabular}{P{0.01cm}P{2.6cm}P{2.6cm}P{2.5cm}P{2.4cm}}
    & Original Image & 600m visibility & 300m visibility &150m visibility
    \end{tabular}
 	 \caption{Image contrast and color quality degradation due to fog and resultant deterioration in the segmentation performance. Row-1: images with decreasing visibility. Row-2: corresponding segmentation outputs. 
 	  Row-3: error maps, where the \textbf{grey} and \textbf{black} regions show correct and erroneous segmentation respectively, while the \textbf{white} regions are untagged. \textcolor{black}{ The segmentation performance deteriorates with increasing fog density. The results are generated using \cite{mlsl2020}, an adaptation approach for GTA to (non-foggy) Cityscapes}.
 	 }
 
 	\label{img:intro}
\end{figure*}

\begin{figure*}[t]
 	\centering
 	\includegraphics[width= \textwidth]{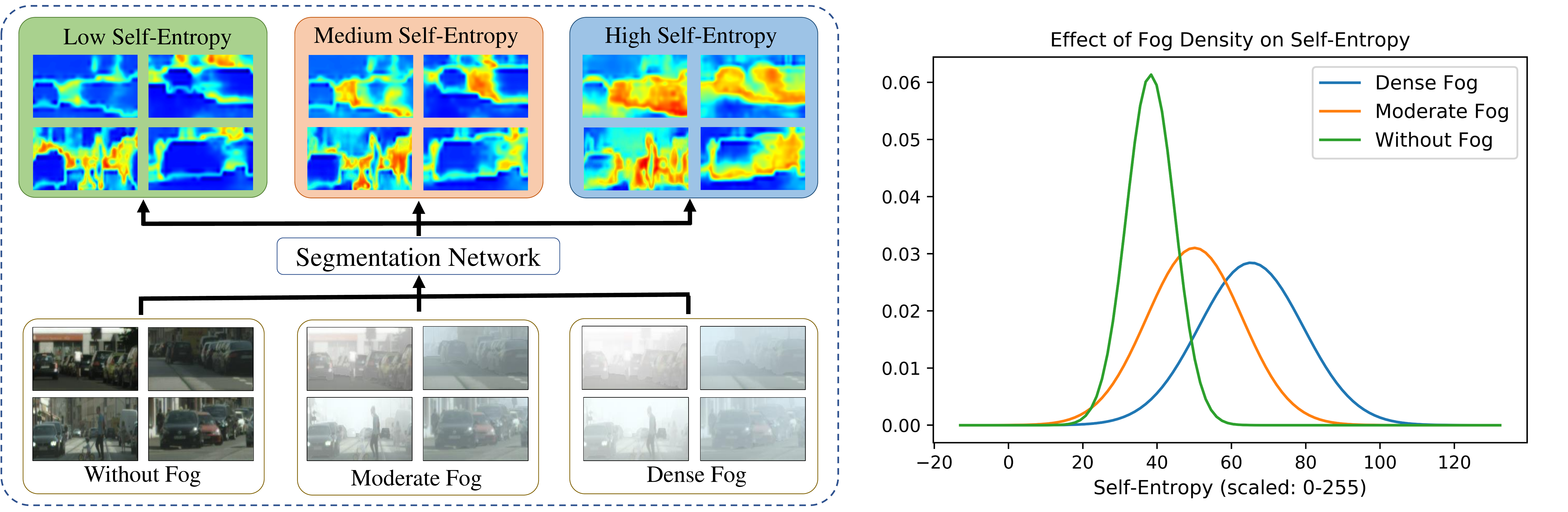}
 	\begin{tabular}{P{5.0cm}P{6.0cm}}
    (a) & (b) 
    \end{tabular}
 	\scriptsize
 	\caption{Relationship between fog density and \textcolor{black}{uncertainty measured by self-entropy} in segmentation probabilities. (a) 
 	\textcolor{black}{Self-entropy maps of semantic segmentation computed over same images as fog changes from none to dense. The denser the fog, the higher the self-entropy.}
 	(b) shows the self-entropy distributions for dense, moderate, and without fog image patches. The mean self-entropy increases with increasing fog density. (The visualizations are generated using a GTA dataset trained model, \textcolor{black}{and for better visualization image-patches are shown instead of full images}.).}
 	\label{img:ent-prob}
\end{figure*}

In this work, we present a novel self-supervised domain adaptation method, \textit{FogAdapt}, for semantic segmentation of images captured in dense foggy weather. 
In foggy conditions, the image contrast and color quality drop significantly degrading the clarity and visibility of the scene. 
This occurs due to the presence of particles in the atmosphere which scatter and absorb light \cite{tan2008visibility, narasimhan2003contrast}. 
\textcolor{black}{Since these particles might be non-uniformly present in different parts of the scene, fog could eventually be of different densities at different locations. Similarly, depending upon the distance between the camera and the objects, fog affects the visibility of objects differently.} 
Specifically, the visibility is decreased with increasing distance, e.g., the farthest objects are more difficult to recognize. 
\textcolor{black}{The source trained segmentation models and even adapted for non-foggy scenarios of a similar scene \cite{wu2019Resnet38,mlsl2020} fails  to mitigate the dense fog, and ends up with deteriorated performance. }
An illustration is shown in  Fig. \ref{img:intro}, where keeping everything else constant, as the fog density is varied from clear to dense, the corresponding semantic segmentation deteriorates accordingly, with the farthest regions most affected.
\textcolor{black}{Combination of these variations results in a considerably large set of scenarios, making collection and labeling costly and laborious, especially for the semantic segmentation and robust supervised learning techniques.}
Hence, a robust unsupervised domain adaptation (UDA) method for such a challenging scenario of dense fog in the target images is needed.


 
\begin{figure*}[htb]
 	\centering
 	\includegraphics[width= \textwidth]{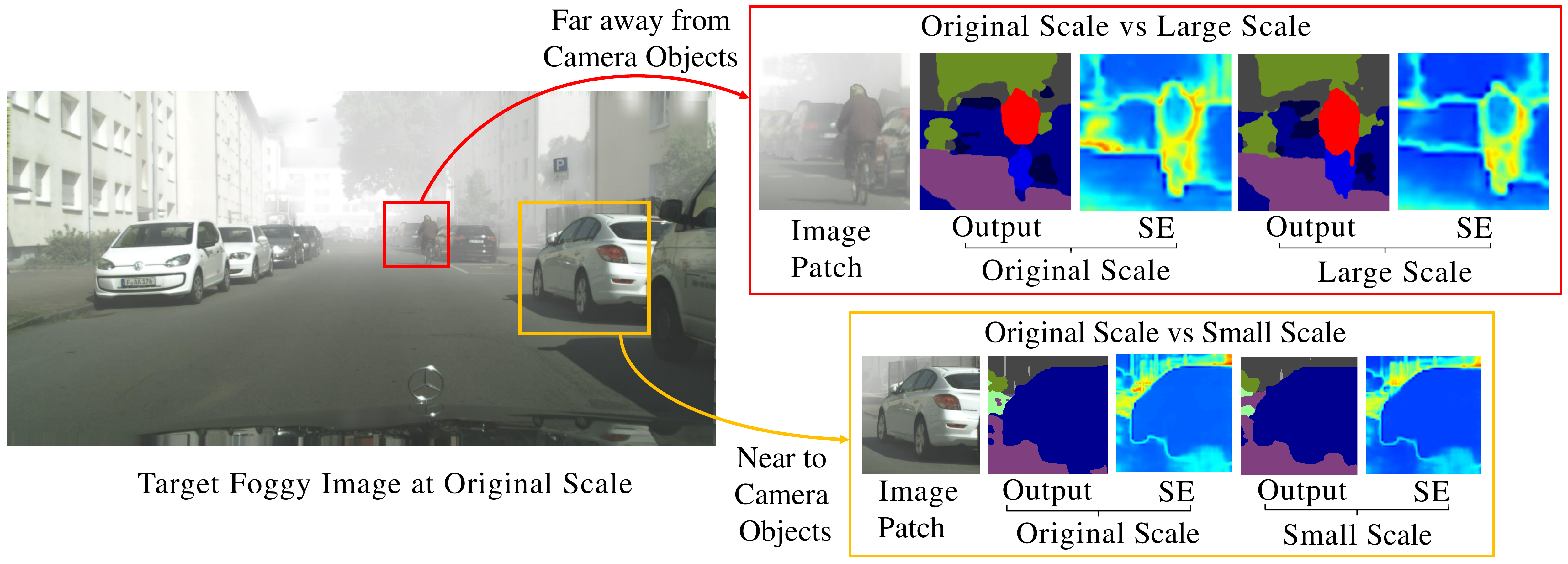}
 	\scriptsize
 	\caption{If an object is far or near the camera, resizing the foggy input image has a different effect on the \textcolor{black}{ self-entropy (SE) map.} 
 	Segmenting foggy scenes at a higher scale provides extra local context, i.e., minimizes the effect of fog by producing better segmentation with comparatively sharp edges. Contrary to that, segmenting images at a lower scale produce better outputs for large and near to camera objects disguised by fog.  \textcolor{black}{These visualizations are generated using source domain (GTA) trained segmentation model.}}

 	\label{img:scale-prob}
\end{figure*}

\textcolor{black}{ To counter the challenges posed by domain shift in dense foggy scenes, we present a novel self-supervised domain adaptation method (FogAdapt) for UDA of foggy scenes segmentation. 
Our domain (fog) specific empirical analysis led us to discover relationships between the effect of fog and road-scene segmentation. 
\textcolor{black}{
We exploit the relationship between \textbf{uncertainty, measured by}  self-entropy, and the density of fog (Fig. \ref{img:ent-prob}) by deﬁning a self-entropy minimization loss for the target images,
when a source (clear weather images) trained semantic segmentation model is utilized for the target (foggy image) dataset.} 
Similarly, we explore the image scale and fog relationship (Fig. \ref{img:scale-prob}) and generate pseudo-labels at pixel level by exploiting the consistency constraint over image scaling to counter the effect of fog. Below we discuss in detail the empirical analysis and the proposed solution specifically designed to counter the effects of fog with further details in Section \ref{sec:method}. }

\textit{Self Entropy loss \& Fog Density:} Images taken in fog exhibit degradation of color quality, low contrast, and other artifacts associated with low visibility. 
Overall this results in images with texture, edges, and color information deteriorated enough to make it difficult to differentiate between different objects and stuff classes. 
This loss of information results in a confused semantic segmentation network (trained in normal weather), making it unable to differentiate between different category pixels, and resulting in high self-entropy. 
This relationship between the self-entropy and the density of the fog has been presented in Fig. \ref{img:ent-prob}. 
To observe how the self-entropy changes with respect to the fog-density, we use Foggy-Cityscapes, a simulated fog added real imagery dataset \cite{sakaridis2018semantic}, where the same images and their foggy versions at multiple visibility ranges are available.
We manually choose and extract small patches ($100 \times 150$ pixels) from the same locations  of the normal images and \textcolor{black}{their modified versions} with simulated dense and moderate fog added to it.  
Self-entropy maps are computed for each patch after passing these images through semantic segmentation network, \textbf{trained on GTA} non-foggy images. Few of them are shown in Fig. \ref{img:ent-prob}.
\textcolor{black}{The three distributions obtained for all the extracted patches at respective fog levels (Fig. \ref{img:ent-prob} (a)), visualized in Fig. \ref{img:ent-prob} (b)}, indicate a strong relationship between fog density and self-entropy, i.e., the denser the fog, the higher the self-entropy. 
\textcolor{black}{This lead us to our hypothesis that minimizing the self-entropy may force the network to learn to compensate for the information loss occurring due to the fog.}


\textit{Scale Invariance \& Fog Density} :
Previously, LSE \cite{LSE_2020_Naseer} introduced scale-invariant examples in the target dataset to minimize the inconsistency between normal and larger scales. More specifically, they observed that in clear weather conditions, images at normal scale are segmented well instead of larger scale and hence they generated pseudo-labels at normal scale. However, in dense foggy scenes, this hypothesis is not completely true. 
\textcolor{black}{In foggy scenes, resizing results in different segmentation accuracy at different locations of the input image depending upon the density of fog and how far or near the object is from the camera.}
\textcolor{black}{This is especially true for the road scenes.}
Due to fog,  the objects that are far from the camera (and hence smaller in scale) have lower visibility, (Eq.\ref{eqn:3gn}) making it further challenging to segment it correctly. 
Since we are employing a self-supervised training approach, pseudo-labels for the small and far away objects disguised in fog will thus not be available as they will have low segmentation scores. 
Therefore, we propose a scale-invariant pseudo-labels generation process for foggy scenes adaptation by exploiting the relationship between scale, fog, and self-entropy (Fig. \ref{img:scale-prob}). 
We make a reasonable assumption that pseudo-labels should be scale-invariant.
Using the same source trained model, the input target image is segmented at multiple image scales (higher and lower spatial resolution than original) independently and the output probability volume is aggregated. 
Segmenting at large scale extrapolates the local context and hence produces better segmentation and low entropy for faraway dense foggy regions compared to normal scale. Similarly, segmenting at small scale benefits large and near to camera objects disguised by fog as shown in Fig. \ref{img:scale-prob}.
The combined effect of these three scales produces better pseudo-labels compared to single normal scale as shown later in Table. \ref{table:ablation-pl}. 

\noindent \textcolor{black}{To summarize, this work produces the following contributions.} 
\begin{enumerate}[noitemsep,nolistsep]
\item A self-supervised domain adaptation strategy for foggy scenes segmentation with pixel-level pseudo-labels to adapt the output space. 
\item Exploiting relationship between the image scale and fog-density to design a strategy for generating \textit{scale invariant} pixel-wise pseudo-labels.
\item Based on empirical evidence, we define a relation between fog density and self-entropy, i.e., self-entropy minimization loss to  mitigate the effects of dense fog in segmentation model and produce confident segmentation output.
\item We show state-of-the-art (SOTA) performances on benchmark datasets by augmenting the scale invariance and self-entropy with \textit{spatial distribution priors} \textcolor{black}{of the source dataset.}
\end{enumerate}

\noindent The reminder of the paper is arranged as follows: Section \ref{sec:relatedWork} describes related work. Section \ref{sec:method} details the proposed approach and Section \ref{sec:exp} presents the experiments and results. In Section \ref{sec:Conclusion} we summarize our work for the conclusion.

\section{Related Work}
\label{sec:relatedWork}


\textcolor{black}{
Domain adaptation approaches have been presented to overcome the domain shift specific to the case of semantic segmentation \cite{tsai2018learning, zou2018unsupervised, vu2019advent, mlsl2020}. However, very little attention has been devoted to address domain shift caused by foggy weather conditions \cite{sakaridis2018model, sakaridis2018semantic, dai2019curriculum}. }
\textcolor{black}{Below, we provide a succinct review of schemes related to generic domain adaptation for Semantic Segmentation followed by the schemes specific to Domain Adaptation for Foggy Scenes Segmentation.}

\subsection{Domain Adaptation for Semantic Segmentation}

\textcolor{black}{
Adversarial learning based UDA of semantic segmentation is the most explored approach in literature \cite{kim2020learning, zhang2018fcan, kim2019bidirectional,dlow_2019_CVPR,chen2017road}. In UDA, adversarial loss-based training is leveraged for input space adaptation (re-weighting) \cite{zhang2018fully, hoffman2017cycada}, feature matching \cite{iqbal2020wan,structure_2019_CVPR, chen2017no,mancini2018boosting, sankaranarayanan2018learning}, structured output matching \cite{tsai2018learning, vu2019advent, kim2019bidirectional} or combination of these strategies \cite{kim2019bidirectional, dada_2019_ICCV, zhang2018fcan}. 
However, due to the global nature of adversarial learning even if the objective is to match the output probabilities or the high dimensional feature representation at latent space, the adversarial domain adaptation alone produces sub-optimal results \cite{vu2019advent, kim2019bidirectional, iqbal2020wan}.}

Besides adversarial learning, self-supervised domain adaptation is gaining attention for many computer vision applications \cite{zou2018unsupervised, mlsl2020, khodabandeh2019robust}. The authors in \cite{zou2018unsupervised, zou2019crst} presented a class balanced pseudo-labels generation and confidence regularized self-training with class spatial priors.
The authors in  MLSL \cite{mlsl2020} leveraged spatial invariance to generate consistent pseudo-labels at pixels and image level for UDA of semantic segmentation in clear-weather scenes.
LSE \cite{LSE_2020_Naseer} tried to generate scale-invariant examples and minimized the loss between pseudo-labels generated at normal scale and its zoomed version. 
\textcolor{black}{Compared to proposed FogAdapt, the LSE \cite{LSE_2020_Naseer} and MLSL \cite{mlsl2020} do not exploit multi-scale information during pseudo-label generation, hence producing inferior performance when exposed to foggy scenes.} 
Zhang et al. \cite{zhang2019curriculum} proposed a curriculum domain adaptation by defining land-mark super-pixels classification based loss at the output while addressing the easy examples first. PyCDA \cite{Lian_2019_pycda} combined \cite{zou2018unsupervised} and \cite{zhang2019curriculum} in a single framework to generate pseudo-labels at multiple sized windows.
The authors in \cite{vu2019advent} used a direct entropy minimization (only high entropy pixels) approach along with the adversarial learning applied on the self-entropy maps of the semantic segmentation. 


\subsection{Domain Adaptation for Foggy Scenes Segmentation}

\subsubsection{Image Defogging/Dehazing}
\textcolor{black}{Color quality and contrast of the outdoor scenes are degraded due to fog/haze. There have been many classical \cite{he2010single, kim2011single, kim2013optimized, ancuti2014effective} and deep learning  \cite{chen2019gated, morales2019feature, golts2019unsupervised, du2018recursive, kim2019bidirectional, liu2019physics} based methods trying to improve color quality or contrast enhancement with an attempt to defog or dehaze. However, as the fog density increases, the defogging models' performance is degraded significantly. Therefore, the attempt to use them as pre-processing step before feeding the data to computer vision models created/trained in normal light settings does not provide desired performance enhancement \cite{pei2018does}.}

\subsubsection{Foggy Scenes Segmentation and Adaptation}
\textcolor{black}{
Besides the great progress for generic semantic segmentation and domain adaptation, very little attention is being devoted to handle foggy scenes. This is mainly due to the unavailability of annotated datasets for foggy scene segmentation. 
The authors in \cite{sakaridis2018semantic} leveraged the stereo property of Cityscapes images to estimate the depth and proposed a fog simulation method for real imagery. They tried to add synthetic fog to real Cityscapes images at multiple fog density levels defined by Eq. \ref{eqn:3gl} to generate synthetic fog added to real images with multiple visibility ranges (Fig. \ref{img:intro}). Alongside, they also developed a small annotated dataset having real foggy scenes; \textit{Foggy Driving}.
Further, they fine-tuned the normal Cityscapes trained model on Foggy-Cityscapes images to address foggy scenes in real imagery. 
Similarly, the authors in \cite{hahner2019semantic} tried to address the real foggy scenes segmentation problem with the help of purely synthetic foggy data. They fine-tuned the normal cityscapes trained models on the synthetic foggy images. 
Sakaridis et al. \cite{sakaridis2018model, dai2019curriculum} proposed a curriculum adaptation learning approach for real foggy scenes understanding. They developed a large dataset, \textit{Foggy Zurich}, by capturing road driving scenes under real foggy scenarios. The dataset is unlabeled except a small chunk of 40 images that have dense annotations available. They adapted to \textit{Foggy Zurich} alongside the fully labeled Foggy-Cityscapes images. They generated pseudo-labels for target images using \cite{sakaridis2018semantic} and defined a fog estimator for curriculum learning.  \textcolor{black}{However, they did not investigate the relationship between the fog density and induced uncertainty for pseudo-labels generation and model adaptation.} }

\textcolor{black}{
In summary, the existing solutions for foggy scenes adaptation have multiple shortcomings. The generic adaptation methods fail to perform in foggy conditions due to lack of domain knowledge. The fog-specific approaches proposed in \cite{sakaridis2018semantic, hahner2019semantic, sakaridis2018model, dai2019curriculum} do not specifically investigate the effect of fog density and respective induced uncertainty. Besides, we introduce a self-supervised domain adaptation approach for foggy scenes by exploiting the relationships between fog density and uncertainty and scale invariance. 
}

\section{Approach}
\label{sec:method}
In this section, we present the details of the proposed approach for self-supervised domain adaptation of semantic segmentation model for dense foggy scenes. 
We start with an introduction to the optical model for fog, basic architecture for semantic segmentation \cite{wu2019Resnet38} and self-training method for domain adaptation \cite{zou2018unsupervised, mlsl2020}. 
Next, we present the proposed FogAdapt algorithm including the loss functions. 

\subsection{Optical Model for Fog}

In general, the image fogging/hazing process is often represented as the physical corruption model given by Eq. \ref{eqn:3gn} \cite{sakaridis2018semantic}
\begin{equation}
  I_d (r, c) = J(r, c) ~ t(r, c) + A(1 - t(r, c)),
\label{eqn:3gn}
\end{equation}
where $I_d$ is degraded image, $t$ is the transmittance map, $J$ is the fog-free radiance of the original image and $A$ represents the global atmospheric lightning ($(r, c)$ in Eq. \ref{eqn:3gn} shows the pixel locations). The transmittance map $t$ is dependent on the distance $l(r, c)$ of the observer from the object having a homogeneous medium and is given by Eq. \ref{eqn:3gl},
\begin{equation}
\centering
  t(r, c) = exp(-\beta~l(r, c)).
\label{eqn:3gl}
\end{equation}
%
The parameter $\beta$ is used to control the density of fog as leveraged by \cite{sakaridis2018semantic}. 
Compared to daylight imagery with clear weather conditions, the foggy scenes are more challenging. 
As highlighted earlier, extensive research has been done on image defogging/dehazing, while less attention is being paid to foggy scenes segmentation and adaptation.

\subsection{Self-Supervised Domain Adaptation: Preliminaries}

\begin{figure*}[!t]
	\centering
	\includegraphics[width= \textwidth]{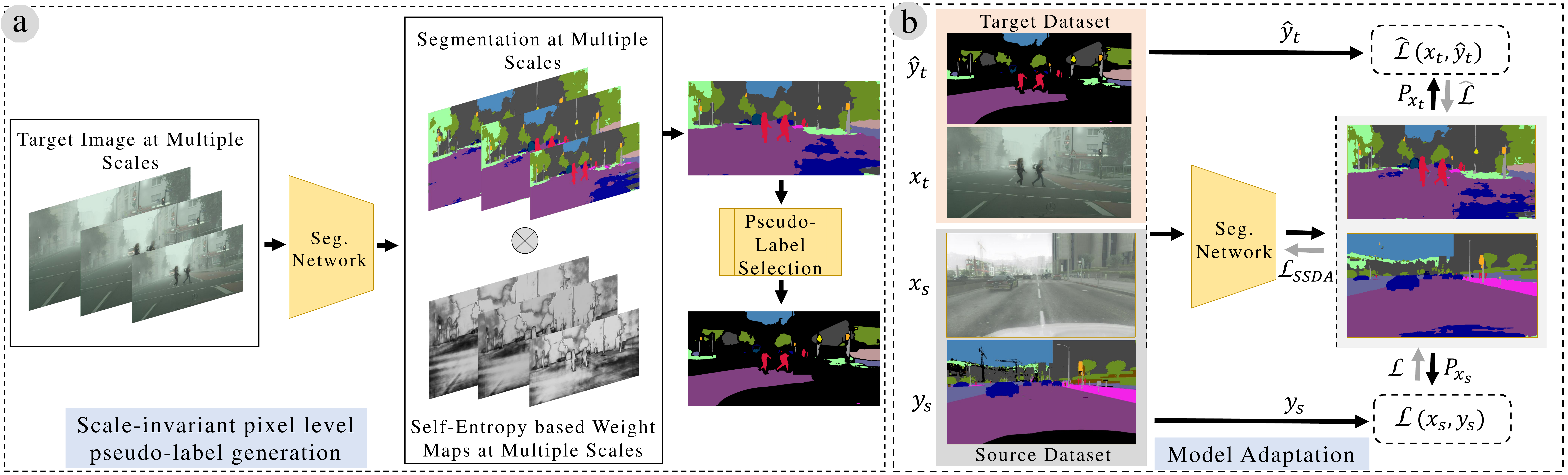}
	\caption{
	The proposed FogAdapt framework. 
	(a) Scale-invariant pseudo-labels generation process where, 1) a target image is resized at multiple scales. 2) the resized versions of the image are segmented independently, 3) uncertainty (self-entropy) based weight maps for each image scale are defined and the outputs are weighted respectively 4) the weighted outputs are then resized to the original scale and recombined, and 5) most confident pixels are assigned pseudo-labels. (b) shows the semantic segmentation adaptation using the generated pseudo-labels for target domain images in (a) and the true labels of source domain images simultaneously.  \textcolor{black}{$x_s, y_s, x_t, \text{and} ~\hat{y_t}$ are source image (after translation), source image labels, target image and target image pseudo-labels, respectively. }
	}
	
	\label{img:proposedModel}
\end{figure*}

Let $\mathrm{X_s}  \subset \mathbb{R} ^{H_s\times W_s\times 3}$ and $\mathrm{X_t} \subset \mathbb{R} ^{H_t\times W_t\times 3}$ be source domain (clear weather) and target domain (foggy) RGB images with spatial resolution $H_s\times W_s$ and $H_t\times W_t$, respectively. 
The true segmentation labels for source domain images are denoted by $\mathrm{y_s} \subset \mathbb{R} ^{H_s\times W_s\times C}$(each pixel location is one-hot encoded) while the ground truth labels for target images are not available. 
$C$ is total number of classes. 
\textcolor{black}{Let $\mathcal{F}$ be the fully convolutional semantic segmentation model, with trainable parameters $\uptheta$.
}
For a given source image $x_s \in \mathrm{X_s}$, let the output segmentation probability volume be denoted by $P_{x_s}$. 
For source domain images, the segmentation model is trained using the cross entropy loss defined in Eq. \ref{eqn:1},
\begin{equation}
\small
     \mathcal{L} (x_s, y_s) = -\sum_{h_s \in H_s} \sum_{w_s \in W_s}\sum_{c\in C} y_s(h_s, w_s, c) ~\log(P_{x_s}(h_s, w_s, c))
\label{eqn:1}
\end{equation}
where $y_s \in \mathrm{Y_s}$ shows the corresponding ground-truth segmentation labels.
%
Since the true labels for target domain images are not present, we use pseudo-labels generated by the source domain trained model for fine-tuning (adaptation).
The corresponding cross entropy loss for the target images is defined in Eq. \ref{eqn:2},
\begin{equation}
\small
     \mathcal{\hat{L}} (x_t, \hat{y}_t) = -\sum_{h_t \in H_t} \sum_{w_t \in W_t}\sum_{c \in C} \rho(h_t,w_t) ~\hat{y}_t(h_t, w_t, c) ~\log(P_{x_t}(h_t, w_t, c))
\label{eqn:2}
\end{equation}
where $\mathcal{\hat{L}} (x_t, \hat{y}_t)$  is self-supervised training loss for target domain images with pseudo-labels $\hat{y}_t$. 
\textcolor{black}{The $\rho^{(H_t,W_t)}$ is a binary map used to compute and backpropagate loss for only those pixels which are assigned pseudo-labels and ignore otherwise. More specifically, for any pixel location, $\rho(h_t,w_t)=1$ if that pixel is assigned a pseudo-label and $\rho(h_t,w_t)=0$ otherwise. }
The pseudo-labels generation and training processes for FogAdapt are shown in Fig. \ref{img:proposedModel} and detailed in Sec. \ref{sec:sisc}. 
%
During target adaptation, the segmentation network is jointly trained using the generated pseudo-labels of the target images and the ground truth labels of source images. 
The corresponding joint loss function for self-supervised domain adaptation (SSDA) is given by 
\begin{equation}
\small
   \min_{\uptheta} \mathcal{L}_{SSDA} ( x_s, y_s, x_t, \hat{y}_t) = \mathcal{L} (x_s, y_s) + \mathcal{\hat{L}} (x_t, \hat{y}_t)
\label{eqn:3}
\end{equation}
The $\mathcal{L}_{SSDA}$ in Eq. \ref{eqn:3} is minimized following a sequential scheme, i.e., fix the segmentation model weight $\uptheta$ to generate pseudo-labels $\hat{y}_t$ for target samples $x_t$, and then use these pseudo-labels to minimize Eq. \ref{eqn:3} with respect to $\uptheta$.
%
%
These steps are repeated for multiple iterations called \textit{rounds}.
The pseudo-labels generation exploiting scale invariance and other constraints are discussed below. 


	

\subsection{Scale Invariant Pseudo-Labels}
\label{sec:sisc}
To adapt the target domain effectively, accurate pseudo-labels are required. 
However, in dense foggy scenes, it is very difficult to generate accurate and consistent pseudo-labels. 
As described in Fig. \ref{img:scale-prob} and explained in Section \ref{intro}, under dense foggy conditions, image regions behave differently at multiple scales. 
Hence, combining multi-scale output information intelligently (Fig. \ref{img:proposedModel}(a)) is more effective compared to any single scale (Table. \ref{table:ablation-pl}). 
\textcolor{black}{Therefore, we present scale-invariant pseudo-labels created by weighted summation of the probability and uncertainty maps across different image scales.}
As discussed in Section \ref{intro} and shown in Fig. \ref{img:intro}, in foggy scenes the visibility of the object is correlated with density of fog present, and the distance between an object and the sensor. 
This combined effect deteriorates the performance of the semantic segmentation model. 
Pixels of the faraway and near to camera objects camouflaged in fog presents very limited information. 
\
More specifically, the model trained on clear weather images skip out small and \textcolor{black}{washed-out} content with respect to available large neighboring content-generating merged or inferior segmentation output. 
\textcolor{black}{Similarly, due to the limited receptive field of the segmentation model 
compared to large and near to camera objects, i.e., bus, truck etc., the performance is effected negatively.}
The segmentation model might be relying on texture or structure information to make the decision. However, that information is corrupted by the presence of fog, resulting in the decreased confidence for these parts and hence fewer pseudo-labels for it.

To minimize the effect of dense fog on far away objects and \textcolor{black}{to overcome the problems associated with the fixed receptive field being unable to provide the complete coverage to larger and near to camera objects,} we generate scale-invariant pseudo-labels.
Instead of generating pseudo-labels at the original (normal) scale that objects exist in an image, we  process target images at multiple scales (spatial resolutions/zoom levels) to generate pseudo-labels, as shown Fig. \ref{img:proposedModel}(a). To assure semantic consistency and scale-invariance, we assume that objects and stuff should be segmented the same irrespective of the scale they are presented.
\textcolor{black}{We evaluate the target image at three scales, e.g., $S = \{1+s^u, 1, 1-s^l\}$, where $s^u$ and $s^l$ are upper and lower scale parameters and are set to $s^u=s^l=0.25$ empirically (Sec. \ref{sec:ablation-sc})}. The target image is resized according to the three scale parameters into three separate images and segmented independently. 
We combine these probability maps based on the confidence of the segmentation model. \textcolor{black}{Specifically, corresponding to each scale, we generate normalized weight maps ($w_{(1-s^l)}, w ~\text{and} ~w_{(1+s^u)}$) based on the self entropy $\mathcal{H}^{(H_t,W_t)}$ (Eq. \ref{eqn:sem}) of the segmentation probabilities corresponding to each scale. This process for $w_{(1-s^l)}$ is shown in Eq. \ref{eqn:scale-weights}.
%
\begin{equation}
\small
   w_{(1-s^l)} = \frac{1- \mathcal{H}^{(H_t,W_t)} (P_{x_{(1-s^l)}})}
   {\sum_{j\in S} {(1- \mathcal{H}^{(H_t,W_t)} (P_{x_{j}}))}}
\label{eqn:scale-weights}
\end{equation}
%
where $j\in S$ represents the image scale.
Similar equations can be written for $w ~\text{and} ~w_{(1+s^u)}$. Using these weight maps, we obtain a weighted summation after resizing to normal scale to form $P_{x_c} \in \mathbb{R} ^{H_t\times W_t\times C}$.
\begin{equation}
\small
   P_{x_c} = \sum_{j\in S}{w_j~.~P_{x_j}}
\label{eqn:3-sc}
\end{equation}
%
where $w_j~\text{with}~(j\in S)$ are the self-entropy based weight maps for respective scales and the (.) operator shows element-wise multiplication.
The higher the self-entropy, the lower the contribution in  $P_{x_c}$ and vice versa. 
The $P_{x_c}$ is used to select pseudo-label based on the confidence score. This process of scale-invariant pseudo-label generation is shown in Fig. \ref{img:proposedModel}(a). 
Hence, the generated pseudo-labels are scale-invariant and quantitatively better compared to single inference (Table. \ref{table:ablation-pl}).
}

To select pixels with high confidence as pseudo-labels and avoid class distribution imbalance problem, we adapt a class balancing and selecting criteria \textcolor{black}{similar to one} used in \cite{zou2019crst}. 
\textcolor{black}{ Similarly, following \cite{zou2018unsupervised, mlsl2020}, we use the spatial priors (SP) of the source domain labels characterized by the occurrences of an object in a specific location across the whole dataset. For each target image, we scale the output probabilities with the spatial-prior mask of each class, and then select the per-pixel high (maximum) probability values from the probability map $P_{x_c}$. }
These high-probability class values over the whole target set are sorted in descending order of confidence and the pixels with high probabilities are selected as pseudo-labels based on the pre-defined selection portion $s_p$.  Initially, $s_p = 15\%$ of the total pixels belonging to any category and is incremented by 5\% in each round. 
The resultant pseudo-labels are class-balanced, consistent, and scale-invariant representative of the whole target dataset. 


\textcolor{black}{
\subsection{Self-Entropy Minimization for Foggy Scenes Adaptation}
\label{sec:sem}
The density of fog has a direct relation with the information contained in an image, e.g., the denser the fog, the lesser the information. 
\textcolor{black}{This is evident from the semantic segmentation results, where dense foggy regions in an image have high self-entropy values (Fig. \ref{img:ent-prob}).}
With increasing fog density, the segmentation model generates under-confident per-pixel predictions making the entropies high. We leverage this relationship between fog density and self-entropy and define a self-entropy minimization loss (\ref{eqn:l-se}) alongside cross-entropy loss (\ref{eqn:3}). 
The underlying idea for self-entropy minimization is to shift the mean self-entropy of dense foggy scenes towards clear images self-entropy mean, as shown in Fig. \ref{img:ent-prob} (b).  
The self-entropy $\mathcal{H}$ for a target image $x_t$ is given by Eq. \ref{eqn:sem},
\begin{equation}
\small
    \mathcal{H}^{(H_t,W_t)} (P_{x_t}) = - \frac{1}{\log (C)} \sum_{c\in C} P_{x_t}^{(H_t,W_t,c)} \log (P_{x_t}^{(H_t,W_t,c)})
\label{eqn:sem}
\end{equation}
where $\mathcal{H}^{(H_t,W_t)} \in [0, 1]$ is the per-pixel standard entropy defined in \cite{shannon1948mathematical}.  
The loss function based on $\mathcal{H}$ for a target image $x_t$ is given by Eq. \ref{eqn:l-se},
\begin{equation}
\small
    \hat{\mathcal{L}}_{se} (P_{x_t}) = \frac{1}{H_t \times W_t} \sum_{H_t,W_t} ~ \mathcal{H}^{(H_t,W_t)} (P_{x_t}^{(H_t,W_t,c)})
\label{eqn:l-se}
\end{equation}
%
During adaptation, we jointly optimize the pseudo-labels based supervised loss $\hat{\mathcal{L}}$ and the unsupervised self-entropy loss $\hat{\mathcal{L}}_{se}$ for an input target image $x_t$. There is a strong resemblance in Eq. \ref{eqn:2} and Eq. \ref{eqn:l-se}, where the former enforces the segmentation model to assign the correct class to an underlined pixel, while the later tries to maximize individuals confidence scores. }

\subsection{Appearance Adaptation}
\textcolor{black}{
As described in Section \ref{sec:relatedWork}, many algorithms tried to exploit the self-training process by labeling the most confident predictions as pseudo-labels. However, it is very important for a pseudo-label to be accurate, consistent, and invariant. 
To generate such confident pseudo-labels the visual appearance of the target and source domain images also play a vital role. For example, a model trained on normal imagery fails to generate accurate pseudo-labels on dense foggy images. 
Hence, an appearance adaptation step is required to help self-supervised learning paradigms to generate consistent pseudo-labels.}

\subsubsection{Image Translation Module}
\textcolor{black}{
In this work, we leveraged the cycle-consistent adversarial learning algorithm (CycleGAN) \cite{hoffman2017cycada} to transform the source domain images to the visual appearance of the unlabeled target domain images. This process is named as Image Translation Module (ITM). The transformed images are nearly similar in visual appearance with target domain images and are used in the domain adaptation process. 
The loss function for the employed CycleGAN is given by Eq.\ref{eqn:3g},
\begin{equation}
\small
\centering
\begin{split}
    \mathcal{L}_{c-Gan} ( x_s, x_t, G_t, G_s, D_t, D_s) = \mathcal{L}_{GAN} (G_t,D_t,  x_s, x_t) \\
    + \mathcal{L}_{GAN} (G_s,D_s,  x_t, x_s) + \mathcal{L}_{cyc} (x_s, G_s(G_t(x_s))) \\
    + \mathcal{L}_{sc} (x_s, G_t(x_s)),
\end{split}
\label{eqn:3g}
\end{equation}
where $G_s$ and $G_t$ represent the generator from target to source and source to target domain respectively. $D_t$ is the discriminator applied to classify between original target domain images and translated target domain images. $D_s$ expedites the same loss for the target to source transformation. Similarly, $\mathcal{L}_{cyc}$ and $\mathcal{L}_{sc}$ losses are applied to maintain the cycle and semantic consistency respectively. The optimization program can be defined as a min-max criterion given in Eq. \ref{eqn:3go},
\begin{equation}
    \underset{G_t, G_s} {min}  ~\underset{D_t, D_s} {max} ~\mathcal{L}_{c-Gan} ( x_s, x_t, G_t, G_s, D_t, D_s).
\label{eqn:3go}
\end{equation}
The transformed source images with available ground truth, when used in domain adaptation helps to select better pseudo-labels and eventually improves the adaptation performance.}
\subsection{Combined Objective Function}

The composite loss function for self-supervision based UDA of foggy scene segmentation is the composition of both Eq. \ref{eqn:3} and Eq. \ref{eqn:l-se}, and is given by,  
\begin{equation}
\small
\mathcal{L}_{cmp} (x_s, y_s, x_t, \hat{y}_t, c) = \mathcal{L}_{SSDA} ( x_s, y_s, x_t, \hat{y}_t) + \hat{\mathcal{L}}_{se} (P_{x_t})
\label{eqn:5}
\end{equation}

Similarly, the combined loss function for ITM augmented with $\mathcal{L}_{cmp}$ is the combination of Eq. \ref{eqn:3g} and Eq. \ref{eqn:5}
and is given by Eq. \ref{eqn:7-itm},
\begin{equation}
\small
\begin{split}
    \mathcal{L}_{ITM-cmp} = \mathcal{L}_{c-Gan} ( x_s, x_t, F_t, F_s, D_t, D_s)
    + \mathcal{L}_{cmp} ( x_s, y_s, x_t, \hat{y}_t, c).
\end{split}
\label{eqn:7-itm}
\end{equation}
To summarize the proposed approach, we train CycleGAN using Eq. \ref{eqn:3go} to translate source images to look like target images. Next, we generate scale-invariant consistent pseudo labels and adapt the baseline model in an iterative manner to minimize the loss function $\mathcal{L}_{cmp}$ defined in Eq. \ref{eqn:5}.  

\section{Experiments and Results}
\label{sec:exp}
\textcolor{black}{This section discusses experimental details and provides the results of our comparison with state-of-the-art techniques.
We list down different configurations and their acronyms in Table. \ref{table:acronyms} for better readability. }
\begin{table}[H]
\footnotesize
\centering
\caption{Different configurations and their acronyms.}
\resizebox{\columnwidth}{!}{
\begin{tabular}{c|c}
\hline
Acronyms & Configuration\\ \hline
FogAdapt & Self-entropy loss + Scale-invariance based pseudo-labels \\
FogAdapt+ & ITM (Image Translation Module) + FogAdapt  \\
SP-FogAdapt+ & Spatial-Priors during pseudo-labels generation + FogAdapt+ \\
\hline
\end{tabular}
}
\label{table:acronyms}
\end{table}

\begin{figure*}[t]
 	\centering
 	\includegraphics[width= \textwidth]{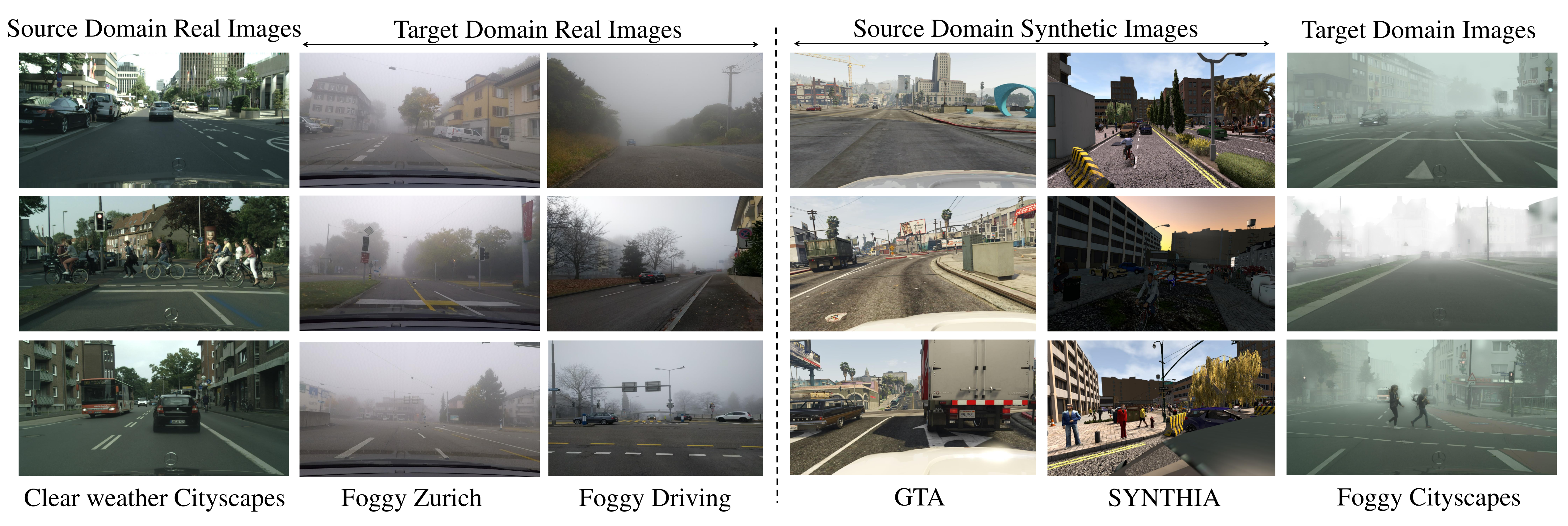}
 	\scriptsize
 	\caption{\textcolor{black}{Sample images from source and target domains. There is a significant difference between the source and target domain images for both the real to real and synthetic to real domain adaptation. }}
 	\label{img:datasets-images}
\end{figure*}

\subsection{Experimental Setup}
\textcolor{black}{
We have performed multiple experiments with various datasets, weather conditions having varying fog densities and settings.}
The key points are discussed below.
\subsubsection{Datasets}
We adapt the standard \textit{real-to-real} and \textit{synthetic-to-real} setup for UDA of foggy scenes segmentation. We use Cityscapes \cite{Cordts2016Cityscapes}, SYNTHIA \cite{Ros_2016_CVPR} and, GTA \cite{Richter_2016_ECCV} datasets as source domain and  \textit{Foggy Driving} \cite{sakaridis2018semantic}, Foggy-Cityscapes \cite{sakaridis2018semantic} and foggy zurich \cite{dai2019curriculum} as real-world target domain datasets.  
\textcolor{black}{Sample images from source and target domain datasets are shown in Figure. \ref{img:datasets-images}. }
\textcolor{black}{
The SYNTHIA-RAND-CITYSCAPES, a sub-set from the SYNTHIA dataset consists of 9400 synthetic frames of spatial resolution $760 \times 1280$. The baseline models and the adapted models are both evaluated with 16 and 13 categories in common between SYNTHIA and Foggy-Cityscapes as described in \cite{vu2019advent} and \cite{zou2018unsupervised} for normal Cityscapes.
Similarly, we use the GTA dataset having 24966 frames with a high spatial resolution of $1052 \times 1914$. Pixel-level labels for classes compatible with Foggy-Cityscapes are available for all 24966 frames.
The Cityscapes dataset consists of 3475 high resolution ($1024 \times 2048$) images with pixel-level annotations, where 2975 images are listed as the training set and the remaining 500 as validation set. Foggy-Cityscapes has the same images as cityscapes with fog being added synthetically by \cite{sakaridis2018model}. 
\textcolor{black}{The \textit{Foggy Driving} (FD) dataset contains 101 images where 33 images have fine annotations and the remaining have coarse labels available. \textit{Foggy Driving Dense} FDD is a subset of the FD dataset having 21 images with very dense fog.}
Similarly, the \textit{Foggy Zurich} dataset contains 3808 high-resolution images of real foggy scenes. However, only a limited set of 40 images is labeled for semantic segmentation.}

\subsubsection{Model Architecture}
\textcolor{black}{
For semantic segmentation of foggy scenes, we use ResNet-38 \cite{wu2019Resnet38} as baseline. ResNet-38 is trained for segmentation of Cityscapes, SYNTHIA, and GTA using the ImageNet pre-trained parameters \cite{russakovsky2015imagenet}.
The architecture of ResNet-38 for segmentation in this work is the same as defined in \cite{wu2019Resnet38, zou2018unsupervised}. 
The Image Translation Module (ITM) is adapted from \cite{hoffman2017cycada}. The ITM is employed to translate source images to the visual appearance of the target domain datasets, e.g., from GTA to Foggy-Cityscapes.}

\begin{table*}[h]
\footnotesize

\caption{Semantic segmentation performance of \textbf{FogAdapt} and its variants compared to SOTA methods on \textit{Foggy Zurich} (FZ), \textit{Foggy Driving-dense } (FDD)  and \textit{Foggy Driving} (FD) test sets when \textbf{adapted from Cityscapes to FZ}.
We present mIoU for all classes compatible with Cityscapes \cite{Cordts2016Cityscapes}, and frequent classes defined for FZ, FDD, and FD respectively.
FogAdapt+: FogAdapt+CycleGAN, \textbf{SP-FogAdapt+}: FogAdapt+ combined with spatial priors defined in \cite{zou2018unsupervised}. The \textbf{bold} text shows highest whereas the {\ul underlined} show the second-highest scores.
}

\centering
\resizebox{0.8\textwidth}{!}{%
\begin{tabular}{p{2.95cm}|P{0.6cm}P{0.6cm}P{0.6cm}|P{0.6cm}P{0.6cm}P{0.6cm}}
\hline
Dataset         & \multicolumn{3}{c}{mIoU over all Classes}          & \multicolumn{3}{|c}{mIoU over frequent Classes}  \\
\hline
Methods         & FZ & FDD & FD  & FZ & FDD & FD  \\ \hline
\hline
AdaptSegNet \cite{tsai2018learning}& 25.0     & 15.8       & 29.7 & -     & -          & -             \\
Semantic \cite{sakaridis2018semantic}& -     & -       & 37.8 & -     & -          & 57.4             \\
Curriculum-FT\cite{dai2019curriculum} & 36.7     & -       & - & 51.7     & -          & -             \\
SUSF \cite{hahner2019semantic} & 42.7 &- & 48.6 & 63.0 & - & 59.5 \\
Model-Ada \cite{sakaridis2018model} & 42.9     & 37.3       & 48.5 & -     & -          & -             \\
MLSL \cite{mlsl2020}  & 45.5     & 43.3       &43.5  & 61.0     & 46.5            & 58.8            \\
Curriculum-Ada \cite{dai2019curriculum} & 46.8     & 43.0       & 49.8 & -     & -          & -             \\ \hline
ResNet-38 (baseline) \cite{wu2019Resnet38} & 33.8     & 39.2       & 39.4 & 48.0     & 43.9          & 56.6             \\ 
Ours(FogAdapt)   & 48.8     & 46.5       & 52.0  & 64.2        & 51.1             & 63.5           \\
Ours(FogAdapt+)   & {\ul 49.8}     & {\ul 47.1}       & {\ul 52.4}  & {\ul 64.4}        & {\ul 51.6}             & {\ul 63.7}            \\
Ours(\textbf{SP-FogAdapt+})    & \textbf{50.6}     & \textbf{49.0}       &\textbf{53.4}  & \textbf{64.6}     & \textbf{53.1}          & \textbf{65.7}            \\ \hline

\end{tabular}%
\label{table:city2zurich}
}
\end{table*}


\subsubsection{Implementation and Training Details}
\textcolor{black}{
To perform the experiments, a core-i5 machine with a single GTX-1080Ti having 11GB of memory is used while MxNet \cite{mxnet15} is used as deep learning framework. 
SGD optimizer with an initial learning rate of $1 \times 10^{-4}$ for the segmentation model and $2 \times 10^{-4}$ for ITM respectively are used for training. 
\textcolor{black}{The scale parameters $s^u$ and $s^l$ are both set to 0.25 (Sec. \ref{sec:ablation-sc}). }
Similarly, $s_p$ is initially set to $ 15\%$ of the total pixels belonging to any category for pseudo-labels selection and is incremented by 5\% in each round (Sec. \ref{sec:sisc}).
Due to GPU memory limitations, we process two images per mini-batch.
The proposed (FogAdapt) iterative process of self-supervised domain adaptation is continued for 4 rounds where each round consists of 2 epochs of training.}

\begin{figure*}[t]
	\centering
	\includegraphics[width= \textwidth]{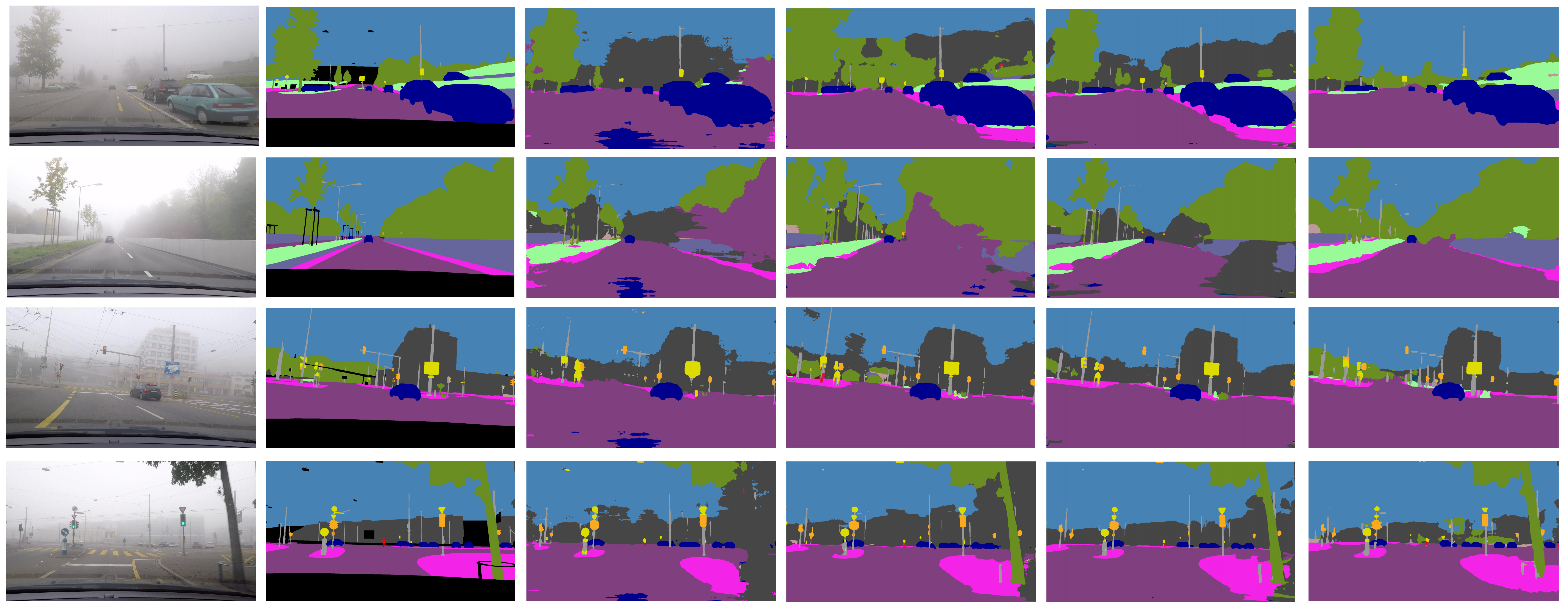}\\
	\footnotesize
	\begin{tabular}{P{1.5cm}P{1.55cm}P{1.7cm}P{1.8cm}P{1.5cm}P{1.5cm}}
    Target Image & Ground Truth & ResNet-38\cite{wu2019Resnet38} &Model-Ada\cite{sakaridis2018model} &Curriculum\cite{dai2019curriculum} & Ours(FogAdapt)
    \end{tabular}
	\caption{Segmentation results on \textit{Foggy Zurich} test set when adapted from Cityscapes. For a fair comparison, we select the images shown by \cite{dai2019curriculum}. The proposed FogAdapt performs better in most of the classes ranging from road to vegetation, train, sky, wall, and buildings.}
	\label{img:city2zurich}
\end{figure*}
\subsection{Experimental Results}
\sloppy
\textcolor{black}{
In this section, we show and discuss the experimental results of \textbf{FogAdapt} compared to ResNet-38 (baseline) and current state-of-the-art (SOTA) UDA approaches for foggy scenes.
Our experiments are two fold: a) In the first setup, we use normal \textit{Cityscapes} (clear-weather images) as source domain and \textit{Foggy Zurich} and \textit{Foggy Driving} (real-foggy imagery) as target domain datasets and (b) in the second setup, we use synthetic datasets, e.g., \textit{SYNTHIA} and \textit{GTA} as source domain and \textit{Foggy-Cityscapes} (synthetic fog added to real images) dataset as target domain.
The performance is reported using a standard evaluation metric for segmentation, i.e.,  Mean Intersection over Union (mIoU). 
The proposed FogAdapt performs superior compared to other domain adaptation approaches with SOTA performance on multiple benchmark datasets varying from synthetic to real for dense foggy conditions.}

\subsubsection{Real Non-foggy to Real-Foggy Scenes Adaptation}
Following \cite{sakaridis2018model, dai2019curriculum}, we adapt the normal (clear-weather) Cityscapes dataset trained source model to the real foggy imagery dataset, \textit{Foggy Zurich}. 
\textcolor{black}{We also evaluate the source only and the \textit{Foggy Zurich} (FZ) adapted models over \textit{Foggy Driving} (FD) and \textit{Foggy Driving-dense } (FDD) test sets (FD and FDD are small sets used for testing only.)} 

\textbf{Cityscapes $\rightarrow$ \textit{Foggy Zurich}: }
Table. \ref{table:city2zurich} summarizes the quantitative results of the proposed approach compared to current SOTA methods. The proposed FogAdapt+ outperforms the ResNet-38 (baseline) and existing approaches with a high margin. Compared to the ResNet-38 \cite{wu2019Resnet38} baseline, we gain $16.0\%$ in mIoU over all classes. We also evaluate FogAdapt over frequent classes in the FZ dataset, e.g., \textit{road, sidewalk, building, wall, fence, pole, traffic-light, traffic-sign, vegetation, sky, and car} defined by \cite{dai2019curriculum}. The FogAdapt+ attains a gain of $22.4\%$ in mIoU over ResNet-38 baseline. 
Similarly, compared to Curriculum-Ada \cite{dai2019curriculum} and Model-Ada \cite{sakaridis2018model}, the proposed FogAdapt+ outperforms them with significant margins of \textcolor{black}{about} $3.0\%$ and $7.0\%$ over all classes respectively. The addition of spatial priors further improves the results as shown by \textbf{SP-FogAdapt+} in Table. \ref{table:city2zurich}. 
\textcolor{black}{We exploit the concept of spatial priors (SP) as presented by CBST \cite{zou2018unsupervised} during pseudo-labels generation. The road scene imagery has a defined spatial structure and the spatial priors are used only if the geometry of the source and target images match. }
To have a fair qualitative comparison, we selected the same images presented in \cite{dai2019curriculum} as shown in Fig. \ref{img:city2zurich}. The proposed FogAdapt shows a significant performance improvement over baselines and existing SOTA methods in most of the categories. 

\textbf{Evaluation on \textit{Foggy Driving}: }
Since the FD dataset is small having 33 images with fine (every pixel is assigned a label)
annotations and the remaining 68 images with coarse (polygon annotations with no clear object boundaries) annotations, this dataset is used only for evaluation as suggested by \cite{sakaridis2018semantic}.  
We evaluate our baseline and FZ adapted models over FD and its' subset FDD. The quantitative results for all 19 classes compatible with FZ dataset are shown in Table. \ref{table:city2zurich}. The proposed FogAdapt+ performs superior compared to baselines and existing SOTA methods. Especially, in case of dense foggy scenes, FDD, our FogAdapt+ achieves a gain of $4.1\%$ and $9.8\%$ in mIoU compared to Curriculum-Ada \cite{dai2019curriculum} and Model-Ada \cite{sakaridis2018model}, respectively. 

Similar to the FZ dataset, we evaluate the FD dataset for frequent classes, e.g., \textit{road, sidewalk, building, pole, traffic-light, traffic-sign, vegetation, sky, person, and car} as defined by \cite{sakaridis2018semantic}. The proposed FogAdapt+ performs significantly better in mIoU, i.e., a minimum gain of $5.1\%$ compared to strong MLSL \cite{mlsl2020}. 

\begin{table*}[h]
\centering
\caption{Segmentation results of adapting GTA to Foggy-Cityscapes. 
The abbreviations "$S_T$", "$A_I$" and "$A_O$" indicates the self-training (self-supervised domain adaptation), input space and output space adversarial learning respectively. PyCDA* is trained with batch size 2 instead of 8 (\cite{Lian_2019_pycda}) due to memory limitations.}
\resizebox{\textwidth}{!}{
\begin{tabular}{l|c|ccccccccccccccccccc|c}
\hline 
\multicolumn{22}{c}{GTA $\rightarrow$ Foggy-Cityscapes}\\
\hline
Methods     & \rot{Appr.} & \rot{Road}  & \rot{Sidewalk} & \rot{Building} & \rot{Wall}  & \rot{Fence} & \rot{Pole}  & \rot{T. Light} & \rot{T. Sign} & \rot{Veg.} & \rot{Terrain} & \rot{Sky}   & \rot{Person} & \rot{Rider} & \rot{Car}   & \rot{Truck} & \rot{Bus}   & \rot{Train} & \rot{M.cycle} & \rot{Bicycle} & \rot{mIoU}  \\ \hline \hline
ResNet-38 \cite{wu2019Resnet38}      & -          & 69.5            & 12.9          & 65.6          & 10.5          & 6.8           & 39.5          & 41.7          & 20.4          & 62.7          & 7.5           & 63.5    & 58.5          & {\ul 31.1}    & 62.3          & 16.3          & 31.9          & 1.4           & 22.0          & 10.8          & 33.4          \\
BDL \cite{kim2019bidirectional}      & $A_O$          & 89.6            & 37.6          & 65.4          & 19.5          & 14.4           & 23.2          & 22.7          & 25.5          & 48.9          & {\ul 35.7}           & 39.7    & 50.7          & 29.8    & 79.1          & \textbf{27.9}          & 32.8          & 0.2           & 18.5          & 30.4          & 36.3          \\

CBST \cite{zou2018unsupervised}        & $S_T$    & 74.6   & 30.3   & 73.2   &  7.0   & 20.0   & 40.7   & 47.4   & 35.4   & 53.0   &  5.8   & 65.4   & 47.7   & 21.7   & 75.4   & 21.7   & \textbf{39.1}   &  5.5   & 18.5   & 33.5   & 37.7 \\

MLSL \cite{mlsl2020} & $S_T$        & 81.5   & 33.6   & 76.6   &  7.9   & 23.1   & 41.1   & 47.5   & 35.9   & 52.0   &  6.1   & 64.9   & 54.1   &
 27.8   & 81.2   & 16.5   & 37.7   &  1.5   & 17.4   & 36.7   & 39.1 \\ 

LSE \cite{LSE_2020_Naseer}    & ST  & 81.9    & 29.0    & 73.3    & 20.7    & 23.2    & 29.1    & 36.9    & 32.7    & 70.4    & 13.4    & 60.9    & 54.8    & \textbf{33.0}    & 75.9    & 25.1    & 27.7    & 7.1    & 25.4    & 33.2    & 39.7 \\ 

PyCDA* \cite{Lian_2019_pycda}    & $S_T$                             & 80.9            & 23.7           & 72.4    & 17.2          & \textbf{28.3}    & 27.1          & 38.4    & 17.6          & {\ul 72.0}    & \textbf{39.9}           & \textbf{74.2}          & \textbf{64.3}          & 24.3          & 72.4          & {\ul 26.2}          & 19.1          & 0.6          & 24.2	        & 36.9          & 40.0    \\ \hline  

Ours (FogAdapt)  & $S_T$  & {\ul 85.1}	  & 31.7	  & {\ul 76.7}	  & 16.5	  & 20.3	  & 41.2	  & 46.2	  & 34.9	  & 70.8	  &  9.1	  & 63.8	  & 53.9	  & 26.2	  & 81.5	  & 22.0	  & {\ul 38.0}	  &  5.9	  & 19.0	  & 36.3	  & 41.0 \\

Ours (FogAdapt+)  & $S_T+A_I$   & 84.1   & {\ul 37.5}   & 76.1   & {\ul 21.9}   & 22.4   & {\ul 41.9}   & {\ul 48.6}   & {\ul 44.2}   & 58.3   &  8.2   & 59.0   & {\ul 58.9}   & 23.0   & \textbf{81.9}   & 26.0   & 33.0   &  {\ul 9.3}   & \textbf{31.1}   & {\ul 40.0}   & 42.8 \\
 
Ours (SP-FogAdapt+)  & $S_T+A_I$   & \textbf{89.6}   & \textbf{41.0}   & \textbf{77.7}   & \textbf{22.6}   & {\ul 24.7}   & \textbf{42.0}   & \textbf{49.9}   & \textbf{47.7}   & \textbf{80.3}   & 16.4   & {\ul 68.7}   & 50.2   &
 21.9   & {\ul 81.6}   & 25.3   & 34.5   & \textbf{10.7}   & {\ul 30.4}   & \textbf{40.5}   & \textbf{45.0} \\ \hline

\end{tabular}
}
\label{table:gta2cityFoggy}
\end{table*}


%
\begin{figure*}[t]
	\centering
	\includegraphics[width= \textwidth]{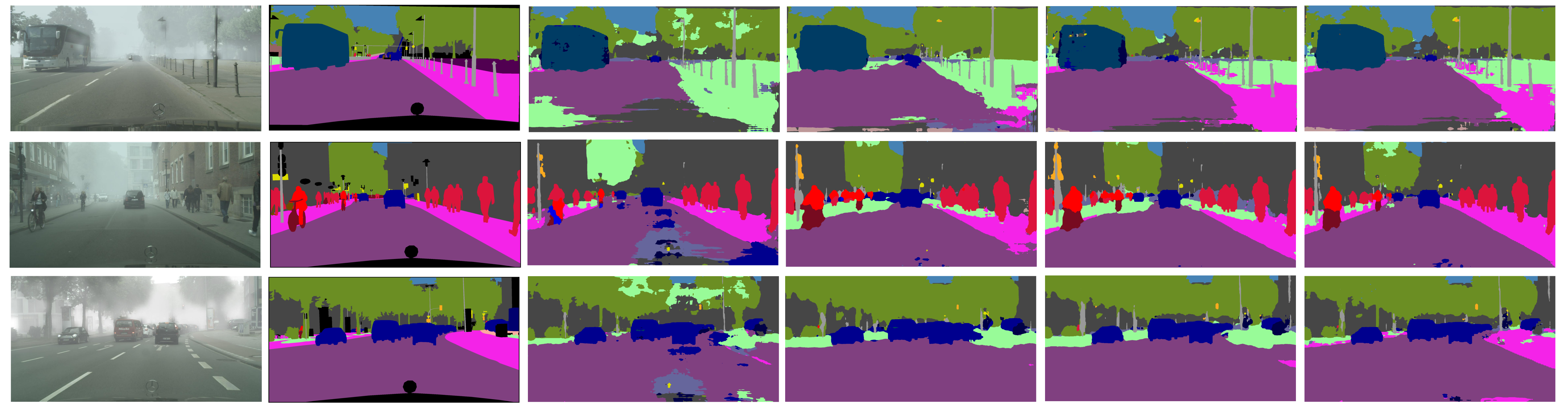}\\
	\footnotesize
	\begin{tabular}{P{1.5cm}P{1.45cm}P{1.8cm}P{1.5cm}P{1.45cm}P{1.4cm}}
    Target Image & Gound Truth & ResNet-38 \cite{wu2019Resnet38} &CBST \cite{zou2018unsupervised} & MLSL \cite{mlsl2020} & Ours(FogAdapt)
    \end{tabular}
    
    \caption{Semantic segmentation qualitative results on the Foggy-Cityscapes validation set when adapted from the GTA dataset trained model. The FogAdapt performs better compared to existing methods. Specifically, the small, thin and far away objects disguised in fog and the stuff classes like road, sidewalk, buildings and sky are segmented better.}
	\label{img:gta2cityFoggy}
\end{figure*}

\subsubsection{Synthetic to Real-Foggy Scenes Adaptation}
To comprehensively test the proposed approach, we also perform synthetic to real foggy domain adaptation experiments. Specifically, we use synthetic datasets, e.g., \textit{SYNTHIA} and \textit{GTA} as source domain datasets and \textit{Foggy-Cityscapes} as target domain. The Foggy-Cityscapes dataset has the real Cityscapes images with synthetic fog added as proposed by \cite{dai2019curriculum}. Foggy-Cityscapes has three levels of fog, e.g., low fog with 600-m visibility, medium fog with 300-m visibility, and dense fog with 150m visibility (Fig. \ref{img:intro}). In this work, we have adapted our models to dense fog scenarios of the Foggy-Cityscapes dataset.   

\textbf{GTA $\rightarrow$ Foggy-Cityscapes: } 
\textcolor{black}{
As indicated in Table. \ref{table:gta2cityFoggy}, the proposed FogAdapt+ for self-supervised domain adaptation shows SOTA performance. 
More specifically, the FogAdapt+ improves the segmentation performance on dense Foggy-Cityscapes by $11.6\%$, $7.3\%$ and $5.9\%$  compared to ResNet-38 baseline and previous SOTA methods, i.e., CBST\cite{zou2018unsupervised}, and MLSL\cite{mlsl2020},  respectively. 
Similarly, compared to LSE\cite{LSE_2020_Naseer} and PyCDA\cite{Lian_2019_pycda}, the proposed approach outperform with a minimum margin of $5.0\%$. 
Fig. \ref{img:gta2cityFoggy} shows semantic segmentation results before and after adaptation.
The proposed approach significantly improves the segmentation performance of dense foggy scenes compared to the source-only model and previous SOTA methods.
}

\begin{figure*}[t]
	\centering
	\includegraphics[width= \textwidth]{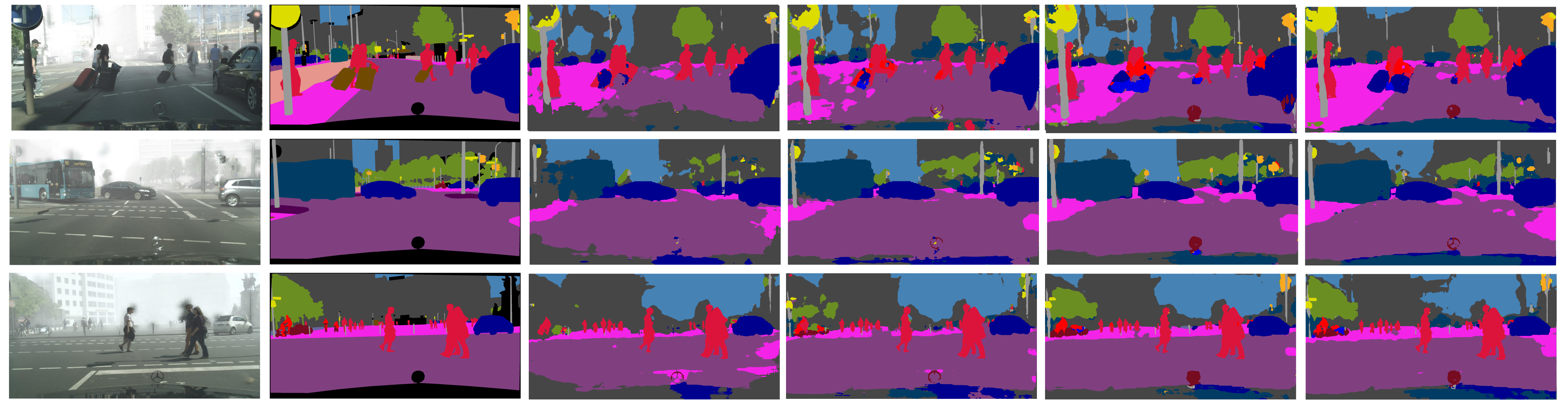}\\
	\footnotesize
	\begin{tabular}{P{1.5cm}P{1.45cm}P{1.8cm}P{1.5cm}P{1.45cm}P{1.4cm}}
    Target Image & Gound Truth & ResNet-38 \cite{wu2019Resnet38} &CBST \cite{zou2018unsupervised} & MLSL \cite{mlsl2020} & Ours(FogAdapt)
    \end{tabular}
    \caption{Qualitative results of semantic segmentation on Foggy-Cityscapes validation set when adapted from SYNTHIA dataset trained model. The proposed FogAdapt performs better compared to \cite{zou2018unsupervised} and \cite{mlsl2020}.}
	\label{img:syn2cityFoggy}
\end{figure*}

\textbf{SYNTHIA $\rightarrow$ Foggy-Cityscapes: } 
\textcolor{black}{
Compared to GTA and FoggyZurich, the SYNTHIA dataset has extra constraints like multiple viewpoint imagery making it a more difficult adaptation task. Table \ref{table:syn2cityFoggy} presents the quantitative results of the proposed approach for 16 overlapping classes between SYNTHIA and Foggy-Cityscapes. Following \cite{ tsai2018learning, clan_2019_CVPR}, we also show the 13 frequent classes results (mIoU*). Compared to baseline ResNet-38, the proposed FogAdapt+ shows a gain of $18.9\%$ and $20.8\%$ in mIoU and mIoU*, respectively.  
Similarly, compared to bidirectional learning \cite{kim2019bidirectional} (output space adversarial learning only), CBST \cite{zou2018unsupervised} and MLSL \cite{mlsl2020}, FogAdapt+ shows a minimum gain of $3.9\%$ and $4.4\%$ in mIoU and mIoU*,  respectively.
A qualitative comparison of FogAdapt with existing SOTA methods is presented in Fig. \ref{img:syn2cityFoggy}
}

\begin{table*}[h]
\centering
\caption{Segmentation results of adapting SYNTHIA to Foggy-Cityscapes. We present mIoU and mIoU* (13-categories as presented by \cite{tsai2018learning}) on the Foggy-Cityscapes validation set.}
\resizebox{\textwidth}{!}{
\begin{tabular}{l|c|cccccccccccccccc|c|c}
\hline 
\multicolumn{19}{c}{SYNTHIA $\rightarrow$ Foggy-Cityscapes}\\
\hline
Methods     & \rot{Appr.} & \rot{Road}  & \rot{Sidewalk} & \rot{Building} & \rot{Wall}  & \rot{Fence} & \rot{Pole}  & \rot{T. Light} & \rot{T. Sign} & \rot{Veg.} &  \rot{Sky}   & \rot{Person} & \rot{Rider} & \rot{Car}    & \rot{Bus}    & \rot{M.cycle} & \rot{Bicycle} & \rot{mIoU} & \rot{mIoU*}  \\ \hline \hline
ResNet-38 \cite{wu2019Resnet38}  & -        & 29.3 & 21.3     & 34.5     & 0.8 & 0.0  & 17.5 & 15.8          & 8.2         & 17.1       & 33.5 & 57.1   & 4.7   & 71.2 & 12.2 & 2.9        & 8.9    & 20.9 & 24.4  \\
BDL \cite{kim2019bidirectional}      & \text{$A_O$}          & \textbf{83.2}            & \textbf{43.2}          & 63.6          & 2.38          & 0.1           & 19.1          & 6.8          & 5.3          & 35.4           & 19.9    & 55.4          & \textbf{ 31.8}    & 65.2              & \textbf{21.0}          & \textbf{27.6}           & 37.1          & 32.3          & 38.1          \\
CBST  \cite{zou2018unsupervised}     & $S_T$   & {\ul 70.5}	     & 31.2	     &  {\ul 57.62}	     & 2.9	     & 0.02	     & 31.7	     & 29.1	     & 23.1	     & 38.5	     & 41.1	     & 61.3	     & 18.9	     & {\ul 75.0}	     & 8.3	     & 11.9	     & 32.1	     & 33.3	     & 38.4\\

MLSL \cite{mlsl2020}   & $S_T$       &48.9	      &27.2	      &53.4	      &11.4	      &0.4	      & {\ul 31.9}	      &\textbf{32.4}	      &21.0	      &49.2	      &40.1	      &\textbf{65.8}	      &24.4	      &\textbf{77.8}	      &{\ul 20.9}	      &19.0	      &\textbf{50.5}	      &35.9	      &40.8 \\ \hline

Ours (FogAdapt)   & $S_T$       & 62.2      & 28.0      & 56.4      &  {\ul 13.1}      &   {\ul 0.7}      & 30.3      &  {\ul 30.1}      & 27.4      &  {\ul 61.7}      & \textbf{61.8}      & 54.9      &  {\ul 30.0}      & 66.1      &  2.6      & 12.1      & 44.8       & 36.4      & 41.4 \\

Ours (FogAdapt+)   & \text{$S_T+A_I$}        & 68.3      &  {\ul 34.0}      & \textbf{64.7}      & \textbf{14.2}      &  \textbf{2.1}      & \textbf{33.2}      & 28.7      & \textbf{31.5}      & \textbf{69.7}      &  {\ul 56.1}      &  {\ul 63.8}      & 29.1      & 66.3      &  6.0      &  {\ul 21.4}      &  {\ul 47.3}      & \textbf{39.8}       & \textbf{45.2} \\
\hline

\end{tabular}
}
\label{table:syn2cityFoggy}
\end{table*}

\subsection{Discussion}
\textcolor{black}{This section investigates the effect of each part of the proposed approach and discusses it in relation to the results obtained.}

\subsubsection{Self-Entropy Minimization}
\textcolor{black}{
As discussed in Sec. \ref{intro}, the higher the fog density, the more uncertain the segmentation model becomes about assigning a class to a specific pixel (Fig. \ref{img:intro}). This decrease in information eventually increases the self-entropy of segmentation probabilities (Fig. \ref{img:ent-prob}). \textcolor{black}{We leverage this relation between entropy and fog density and add an entropy minimization constraint (Eq. \ref{eqn:l-se}) to the total loss function}. Adding this constraint increases the mIoU performance compared to simple pseudo-labels based self-supervised domain adaptation (SSDA) by $3.3\%$ for GTA to Foggy-Cityscapes as shown in Table. \ref{table:ablation}. 
}

\begin{table}[H]
\footnotesize
\centering
\caption{Effect of self-entropy and scale invariance. SSDA: Self-supervised domain adaptation. Here SE is Self-entropy while SI is Scale Invariance.}
\resizebox{\columnwidth}{!}{
\begin{tabular}{c|ccccc}
\hline
\multicolumn{6}{c}{GTA $\rightarrow$ Foggy-Cityscapes} \\
\hline
Methods & ResNet-38 & SSDA & SSDA-SE & SSDA-SI  & FogAdapt \\ \hline
mIoU & 33.4  & 37.0   & 40.3    & 40.2 & 41.0   \\
\hline
\end{tabular}
}
\label{table:ablation}
\end{table}

\subsubsection{Scale-Invariance}
\textcolor{black}{
As the effect of fog increases with the distance between the object and the observer, the scale of an object has a major role (Fig. \ref{img:scale-prob}) in properly segmenting the object. 
\textcolor{black}{Increasing the object's size by resizing the image to a larger scale makes the object under fog clearer as it increases the local contextual information (Fig. \ref{img:scale-prob}). }
This helps in generating comparatively better pseudo labels (Table. \ref{table:ablation-pl}).  
On the other hand, resizing image to a smaller size allows the segmentation algorithm to properly segment near to camera objects disguised by fog which were erroneous previously due to limited receptive field, and capture a more global view. 
\textcolor{black}{As described in Sec. \ref{sec:sisc} , combining these higher and lower scales results in the generation of robust and consistent pseudo-labels (Table. \ref{table:ablation-pl}), which eventually increases the performance over foggy scenes.}
Experimental results show a gain of $3.2\%$ in mIoU compared to single scale for GTA to Foggy-Cityscapes adaptation (Table. \ref{table:ablation}). Similarly, for normal Cityscapes to FZ dataset adaptation, the SI performs superior when combined with SE for FZ, FDD, and FD compared to all previous approaches as shown in Table. \ref{table:city2zurich} (FogAdapt+)).
}

\begin{table}[H]
\footnotesize
\centering
\caption{Effect of incorporating uncertainty weighted scale invariance on the quality of pseudo-labels. Here wSI is Weighted Scale-Invariance.}
\resizebox{\columnwidth}{!}{
\begin{tabular}{c|cc|cc}
\hline
\multicolumn{5}{c}{GTA $\rightarrow$ Foggy-Cityscapes} \\ \hline
&\multicolumn{2}{c}{Start of round 0} & \multicolumn{2}{c}{Start of round 1} \\
\hline
Methods & Normal Scale &  wSI based multi-scale & Normal Scale &  wSI based multi-scale \\ \hline
mIoU & 69.6  & \textbf{71.5} & 71.3  & \textbf{74.3}   \\
\hline
\end{tabular}
}
\label{table:ablation-pl}
\end{table}

\subsubsection{Effect of Input Space Adaptation}
\textcolor{black}{
With ResNet-38 \cite{wu2019Resnet38} as baseline model, we investigate the effect of image translation at input space.
We train CycleGAN \cite{hoffman2017cycada} to translate source images (non-foggy) to the visual appearance (foggy) of the target domain images; GTA/SYNTHIA to Foggy-Cityscapes and Cityscapes to \textit{Foggy Zurich}, respectively. This process generates foggy imagery for the source datasets helping FogAdapt to generate better pseudo-labels. 
Adding this image translation module to FogAdapt: FogAdapt+, significantly improves the segmentation performance for foggy scenes. For GTA to Foggy-Cityscapes, FogAdapt+ gains $1.8\%$ in mIoU as shown in Table \ref{table:gta2cityFoggy}. 
Similarly, for normal Cityscapes to real foggy datasets adaptation the FogAdapt+ gain $1.0\%$, $0.5\%$ and $0.4\%$ in mIoU for FZ, FDD and FD over the FogAdapt respectively (Table. \ref{table:city2zurich}). 
Thus, the input space adaptation has an impact on the pseudo-label generation and adaptation process.}

\textcolor{black}{ The proposed FogAdapt is a complementary method and any image translation approach can be used with it. A better image translation model shall lead to better adaptation performance. To validate this, we use CUT \cite{park2020cut} a more recent approach compared to CycleGAN \cite{hoffman2017cycada} to translate source images to visual appearance of the target domain. This image translation approach improves the mIoU of the SP-FogAdapt+ to 46.0\% compared to 45.0\% previously for GTA to Foggy Cityscapes adaptation.
}


\subsubsection{Effect of Dehazing/Defogging}
\textcolor{black}{
To investigate the effect of the defogging process, we conduct experiments with defogging as input space adaptation. We define an image defogging module (IDM) based on the method proposed by \cite{chen2019gated}. FogAdapt is applied on these defogged images for domain adaptation. 
However, it is observed that defogging methods perform inferior in the presence of dense fog. The same observation was previously reported by \cite{dai2019curriculum}. Compared to defogging as input space adaptation, the source images translation to the target domain appearance performs well (Table. \ref{table:6}).}
\begin{table}[H]
\footnotesize
\centering
\caption{A comparative analysis of image transformation methods in dense foggy scenes adaptation process.}
\resizebox{\columnwidth}{!}{
\begin{tabular}{c|cccc}
\hline
\multicolumn{5}{c}{GTA $\rightarrow$ Foggy-Cityscapes} \\
\hline
Methods & ResNet-38 & FogAdapt & IDM-FogAdapt  & ITM-FogAdapt \\ \hline
mIoU & 33.4  & 41.0   & 40.0    & 42.8   \\
\hline
\end{tabular}
}
\label{table:6}
\vspace{-0.3cm}
\end{table}

\textcolor{black}{
\subsubsection{Effect of Scaling Parameters $s^u, s^l$}
\label{sec:ablation-sc}
We investigate the effect of image scaling and its effect on the quality of pseudo-labels. Table \ref{table:ablation-sc} shows the mIoU of generated pseudo-labels for different upper and lower scaling factors $s^u, s^l$. It is observed that, that very large and very small image scales reduces the pseudo-labels quality significantly. The selected values ($s^u, s^l = 0.25$) of scaling factors produces better pseudo-labels.
%
\begin{table}[H]
\footnotesize
\centering
\caption{\textcolor{black}{Effect of lower and upper scale parameters ($s^l, s^u$) on pseudo-labels.}}
{\color{black}\begin{tabular}{c|cccc}
\hline
\multicolumn{5}{c}{GTA $\rightarrow$ Foggy-Cityscapes} \\
\hline
$s^l, s^u$& (0.5 0.5)  & (0.5, 0.25)   & (0.25, 0.25)    & (0.25, 0.5)   \\
\hline
mIoU& 67.8  & 69.0   & \textbf{71.5}    & 67.9   \\
\hline
\end{tabular}}
\label{table:ablation-sc}
\end{table}
%
%
}

\section{Conclusion}
\label{sec:Conclusion}
\textcolor{black}{
In this paper, we have proposed a self-supervised domain adaptation strategy with self-entropy and scale-invariance constraints for UDA of foggy scene semantic segmentation. 
We empirically establish a relationship between the fog density and self-entropy of the source model's prediction over the foggy images. We exploit this relationship to define a self-entropy minimization objective function to  adapt on images where color quality and contrast has been degraded due to fog. 
Having a fair assumption that under foggy conditions labels of stuff and objects should be the same regardless of their scale, we generate scale-invariant pixel level pseudo-labels.
Scale-invariance helps us to counter the phenomena that in foggy weather, objects farther away are less visible and hence suffer from more information loss. 
The scale invariant pseudo-label generation and the self-entropy minimization for self-supervised domain adaptation allows the segmentation model to learn domain independent features to mitigate the effect of fog density.
Rigorous experiments demonstrate that the proposed self-supervised domain adaptation method augmented with image translation module (ITM) outperforms the existing SOTA algorithms on benchmark datasets: mIoU improves from $46.8$ to $50.6$ on \textit{Foggy Zurich}, $43.0$ to $49.0$ on \textit{Foggy Driving-dense } and $49.8$ to $53.4$ \textit{Foggy Driving} when adapted from Cityscapes to \textit{Foggy Zurich}. 
Effectiveness of self-entropy minimization, scale invariant pseudo-labels, and ITM is highlighted by the considerable improvement of mIoU over the baseline model and SOTA methods. 
\textcolor{black}{Being complementary in nature, when used with state of the art image translation model our results improve to 46.0. for GTA to Foggy-Cityscapes adaptation.}}



\bibliography{bibfile}

\end{document}